\documentclass[letterpaper, 10 pt, conference]{ieeeconf}  % Comment this line out if you need a4paper

\IEEEoverridecommandlockouts                              % This command is only needed if 
\usepackage{amsmath} 
\usepackage{amssymb} 

\usepackage{bm}
\usepackage{amsfonts}
\usepackage{graphicx}
\usepackage{multirow}
\usepackage{booktabs} % To thicken table lines
\usepackage{algorithm}[H]
\usepackage[noend]{algpseudocode}%

\usepackage{booktabs}
\usepackage{threeparttable}
\let\labelindent\relax
\usepackage{enumitem}
\usepackage{gensymb}

\usepackage{hyperref}

\newcommand{\tgparams}{\textbf{x}}

\newcommand{\freq}{$f$}

\newcommand{\algoname}{EETG}
\newcommand{\algonamefull}{Evolved Environmental Trajectory Generators}

\newcommand{\algovarone}{EETG Itr.}
\newcommand{\algovartwo}{EETG Itr.+ Policy}

\newcommand{\pmtgvone}{PMTG (Enc.)}
\newcommand{\pmtgvtwo}{PMTG (Ind.)}

\newcommand{\cmaespmtgvone}{CMAES-PMTG (Enc.)}
\newcommand{\cmaespmtgvtwo}{CMAES-PMTG (Ind.)}

\title{\LARGE \bf
Efficient Learning of Locomotion Skills through the Discovery of Diverse Environmental Trajectory Generator Priors
}

\author{Shikha Surana*$^{1}$, Bryan Lim*$^{1}$, Antoine Cully$^{1}$% <-this % stops a space
\thanks{*Equal Contribution}%
\thanks{$^{1}$Imperial College London, United Kingdom.
{\tt\small \{ss5721, bwl116, a.cully\}@ic.ac.uk}}%
}

\begin{document}

\maketitle
\thispagestyle{empty}
\pagestyle{empty}

\begin{figure*}[h!]
\centering
	\includegraphics[width=1\textwidth]{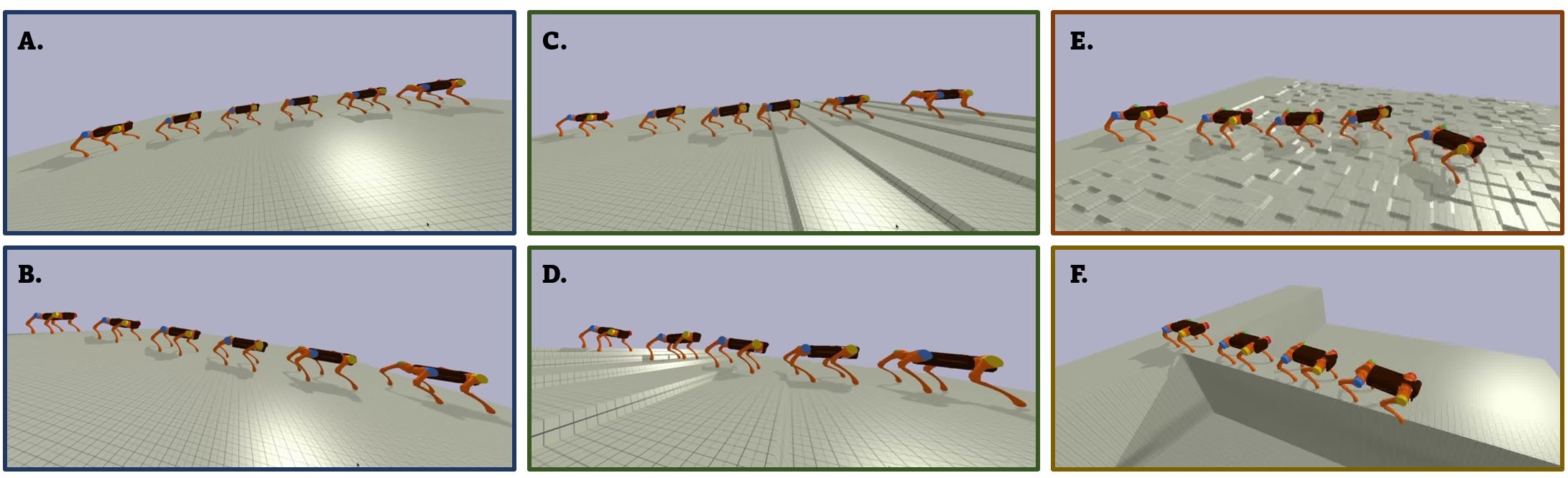}
\centering
\caption{\textbf{A1 quadruped robot, trained with the presented approach, in a variety of challenging environments.} \textbf{(A)} Ascending slope. \textbf{(B)} Descending slope. \textbf{(C)} Ascending stairs. \textbf{(D)} Descending stairs. \textbf{(E)} Uneven/rough terrain. \textbf{(F)} Narrow balance beam.}
\label{fig:feature_fig}
\vspace{-4mm}
\end{figure*} 

%%%%%%%%%%%%%%%%%%%%%%%%%%%%%%%%%%%%%%%%%%%%%%%%%%%%%%%%%%%%%%%%%%%%%%%%%%%%%%%%
\begin{abstract}
Data-driven learning based methods have recently been particularly successful at learning robust locomotion controllers for a variety of unstructured terrains. 
Prior work has shown that incorporating good locomotion priors in the form of trajectory generators (TGs) is effective at efficiently learning complex locomotion skills. 
However, defining a good, single TG as tasks/environments become increasingly more complex remains a challenging problem as it requires extensive tuning and risks reducing the effectiveness of the prior.
In this paper, we present \algonamefull~(\algoname), a method that learns a diverse set of specialised locomotion priors using Quality-Diversity algorithms while maintaining a single policy within the Policies Modulating TG (PMTG) architecture.
The results demonstrate that \algoname~enables a quadruped robot to successfully traverse a wide range of environments, such as slopes, stairs, rough terrain, and balance beams. 
Our experiments show that learning a diverse set of specialized TG priors is significantly (5 times) more efficient than using a single, fixed prior when dealing with a wide range of environments.

\end{abstract}

%%%%%%%%%%%%%%%%%%%%%%%%%%%%%%%%%%%%%%%%%%%%%%%%%%%%%%%%%%%%%%%%%%%%%%%%%%%%%%%%
\section{INTRODUCTION}
Legged robots~\cite{bledt2018cheetah, hutter2016anymal, katz2019mini} have tremendous potential for societal impact as they can be used for applications involving a wide range of environments such as rough, cluttered and unstructured terrain. 
From search and rescue and inspection work~\cite{gehring2021anymal} to carrying heavy payloads, legged robots have the potential to undertake many of the physical activities humans and animals are capable of that are dangerous and unhealthy.

However, legged robots are also underactuated high-dimensional systems with many constraints making them challenging to control.
Recently, reinforcement learning (RL) approaches~\cite{hwangbo2019learning, lee2020learning, siekmann2021sim, rudin2022learning, miki2022learning, margolis2022rapid} have become competitive to more conventional model-based optimization methods~\cite{di2018dynamic, winkler2018gait, kim2019highly, bledt2020extracting}, demonstrating state-of-art locomotion abilities both in simulation and in the real-world~\cite{hwangbo2019learning, lee2020learning}. 
These learnt controllers are especially robust when evaluated across many different environments and perturbations.
Despite these significant advances, learning based approaches in robotics are notoriously known for being sample inefficient and usually requires a large amount of data~\cite{akkaya2019solving, levine2018learning}.
Researchers have tried to address this problem in a number of different ways. For example, improving the sample efficiency of the underlying RL algorithm used~\cite{smith2022legged} or using fast, highly parallel simulators~\cite{makoviychuk2021isaac, rudin2022learning}.
Another effective way is to incorporate useful priors in the learning process.

% Talk/focus in on PMTG
Policies Modulating Trajectory Generator (PMTG)~\cite{iscen2018policies} is one such method which incorporates a parameterized Trajectory Generator (TG) as a prior, separate to the learnt policy. PMTG makes learning complex locomotion tasks easier and demonstrates that a good locomotion prior can significantly help the efficiency of reinforcement learning (RL) methods~\cite{iscen2018policies}.
Lee et al.~\cite{lee2020learning} also used this PMTG framework when demonstrating state-of-the-art locomotion across wide range of environments in the real-world, further demonstrating the effectiveness of the TG prior for locomotion.

However, there are still some questions surrounding the PMTG method. How are the parameters of the TG defined? What parameters make a good prior? The parameters of the TGs used in prior work are usually defined manually by engineers based on intuition of the locomotion task of interest. For instance, a forward swinging TG motion is useful when learning to walk forward~\cite{iscen2018policies}. On the other hand, for more complex tasks such as learning to walk across a diversity of difficult environments, a more generic and unbiased TG motion of stepping up and down in place had to be used~\cite{lee2020learning}.
While this TG choice proved effective, this could reduce the effectiveness of the prior in helping learning and could indicate that the policy still has to do the bulk of the work as it has to deal with the different environments.
For example, a good TG prior for ascending steps would differ from that of descending steps.
In this paper, we address this by learning good priors for tasks instead of manually crafting them. 
More importantly, we also learn a diverse set of specialized priors using Quality-Diversity (QD) algorithms rather than using just a single prior.

The main contribution of our work is a novel framework, \algonamefull~(\algoname)~ for discovering a diverse set of specialized Trajectory Generators (TGs) which act as priors for more efficient learning. 
We demonstrate in our experiments that our method enables a simulated A1 quadruped robot to learn dynamic locomotion behaviors over diverse environment types such as slopes, uneven terrain, and steps.
Our experiments show that \algoname~is as good as learning individual TGs and policies across all environments while being significantly more efficient.
Our work demonstrates that learning a diverse set of TG prior is more effective than a single fixed TG especially when dealing with many tasks and environments.

\section{RELATED WORK}
\textbf{Legged Locomotion.}
Locomotion controllers have traditionally been designed using a modular control framework. 
This framework breaks down the difficult control problem into smaller sub-problems. Each sub-problem makes approximations such as mass-less limbs and point mass dynamics~\cite{di2018dynamic, kim2019highly} and apply heuristics~\cite{bledt2020extracting} which are used alongside trajectory optimization, footstep planning, model predictive control (MPC) methods~\cite{kalakrishnan2010fast, di2018dynamic, winkler2018gait}.

Alternatively, data-driven learning based approaches are fast becoming a go-to option due to recent advances showing impressive robustness and performance~\cite{hwangbo2019learning, lee2020learning, miki2022learning, margolis2022rapid}, while at the same time requiring less modeling and expert optimization knowledge.
One of the key ideas that make learning based methods perform so well and give these controllers incredible robustness especially in the real world, is domain randomization (DR). 
DR of simulator physics and visual features enabled learning dexterous manipulation to solve a Rubik's cube~\cite{akkaya2019solving}.
Hwangbo et al.~\cite{hwangbo2019learning} then demonstrated DR was also effective for locomotion in dynamic legged systems.
This idea of DR can be also extended towards curriculum learning based methods and used beyond just physical parameters and dynamics of the robot. Multiple separate works~\cite{lee2020learning, kumar2021rma, paglieri2021} show that learning through a curriculum of diverse environment types and terrains can result in learnt skills which can be robust and generalize to new environments.
Similarly, our work utilizes environment diversity as strategy to learn robust and generalizable controllers.

As mentioned in the introduction, we also build on the PMTG control architecture~\cite{iscen2018policies} to utilize the effectiveness of priors for learning. 
However, instead of manually crafting a single fixed parameter vector of the trajectory generator (TG), we learn a diverse and high-performing set of specialised TGs.

\textbf{Quality-Diversity.}
Quality-Diversity (QD)~\cite{pugh2016quality, cully2017quality, chatzilygeroudis2021quality} is a growing branch of optimization methods which aims to find a large set of diverse and locally optimal solution. This is in contrast to conventional optimization algorithms which find a single objective maximising solution. QD algorithms have demonstrated their effectiveness across a range of applications including robotics~\cite{cully2015robots, salehi2022few, lim2022learning}, reinforcement learning~\cite{ecoffet2021first, wang2019poet}, video-game design~\cite{fontaine2021illuminating, earle2022illuminating}, engineering design optimization~\cite{gaier2018data} and more.

In the context of robotics, QD algorithms have commonly been used to learn a diverse repertoire of primitive controllers~\cite{cully2013behavioral} which can then be used effectively for downstream applications such a damage recovery~\cite{cully2015robots, chatzilygeroudis2018reset} or with planning algorithms to perform longer horizon tasks such as navigation to a goal~\cite{chatzilygeroudis2018reset, kaushik2020adaptive, lim2022dynamics}. 
In our work, we learn the parameters of a controller which is represented as an open loop TG.
However, a critical difference in our method is that the descriptors/cells of the QD algorithm are not defined by the behavior of the controller but by the environment (i.e. variations of stairs, rough terrain etc.) in which the controller is evaluated in and is selected before evaluation.
This is akin to multi-task MAP-Elites~\cite{mouret2020quality}.

QD-like algorithms like POET~\cite{wang2019poet, wang2020enhanced} have also been used to evolve environments of increasing complexity while learning specialised paired policies for each environment. Similarly, our work also maintains a diversity of environments while trying to discover specialized parameters for each environment. However, we learn specialised TGs instead of specialised policy networks for each environment. We then later learn a single policy over all the specialized TGs. Our method also differs in that we do not evolve environments but define them beforehand as we are interested in discovering specialized priors and not in auto-curricula of the environment. However, we expect this to further help despite requiring additional complexity of an additional optimization process and more compute. We leave this for future work.

\begin{figure*}[t]
\centering
	\includegraphics[width=1.0\textwidth]{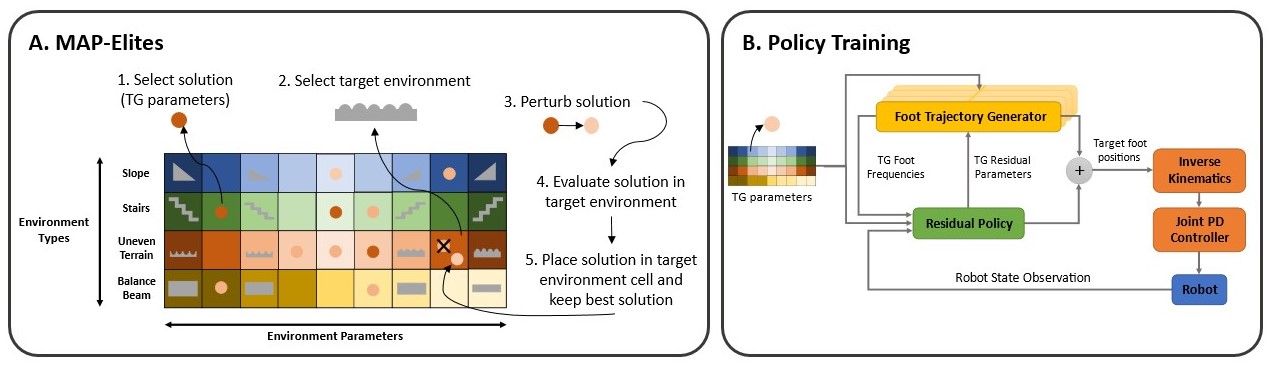}
\centering
\caption{\textbf{\algoname~Overview.} \textbf{(A)} A diverse and high-performing set of environmental priors is learnt using QD optimization.. 
First, a solution and target environment is selected from the grid (1-2). 
The solution's parameters are perturbed (3) and evaluated in the target environment (4). Finally, the solution is placed in the target environment cell, where the best solution in the cell is kept (5). 
\textbf{(B)} Policy optimisation. At each training episode, a target environment and its corresponding prior (i.e., TG parameters) are selected from the grid. 
%The trajectory generator (TG) is parameterised with the TG parameters, 
The residual policy is optimised to modulate the TG actions with ARS~\cite{mania2018simple}.}
\label{fig:method_framework}
\vspace{-4mm}
\end{figure*} 

\section{METHODS}
Our algorithm is composed of two parts: 1) Learning of a diverse set of environment specialized TG priors (Section \ref{method_A})and 2) policy optimization within the PMTG architecture (Section \ref{method_B}).
Figure~\ref{fig:method_framework} shows an overview of the two phases of the \algoname~algorithm. We describe these two parts in the following subsections.

\subsection{Discovering Diverse Specialised Trajectory Generators} \label{method_A}
Our goal is to find a diverse set of priors in the form of TGs, each of which are specialised and high-performing for the corresponding task and environment. 
This problem can be formalised as a Quality-Diversity (QD) optimization problem.
In this subsection, we first define the TG representation used and how the TGs are parameterised. The parameter space of the TGs is then used as the solution space which the QD algorithm searches over.
We then go through our method, \algoname~which applies some key modifications to the original MAP-Elites algorithm, a popular QD algorithm.

\textbf{TG representation and parameterization.} 
In our work, the TG generates an open-loop smooth nominal periodic trajectory for each foot similar to Lee et al.~\cite{lee2020learning}. The trajectory for each foot is defined in Eqn~\ref{eg:ftg}.
\begin{equation}
    \begin{gathered}
        \phi_i = \phi_{i,0} + (\freq_0 + \freq_i)\tau \\ 
        \beta_1 = \frac{\sin(\phi_i + \frac{\pi}{2}) - 1}{2} 
        \hspace{0.7cm}
        \beta_{2} = \frac{\sin(2\phi_i + \frac{\pi}{2}) - 1}{2} \\
        x_i = (s \times \beta_1) + s \\
        y_i =  
        \begin{cases} 
          y_{\delta, i} & \phi_i \in [0, \pi] \\
          y_{\delta, i} - (t \times \beta_2) & \phi_i \in [\pi, 2\pi] \\
       \end{cases}
        \\
        z_i =  
        \begin{cases} 
          h & \phi_i \in [0, \pi] \\
          h - (l \times \beta_2) & \phi_i \in [\pi, 2\pi] \\
       \end{cases}
        \\
    \end{gathered}
    \label{eg:ftg}
\end{equation}

\begin{figure}[t]
\centering
	\includegraphics[width=0.45\textwidth]{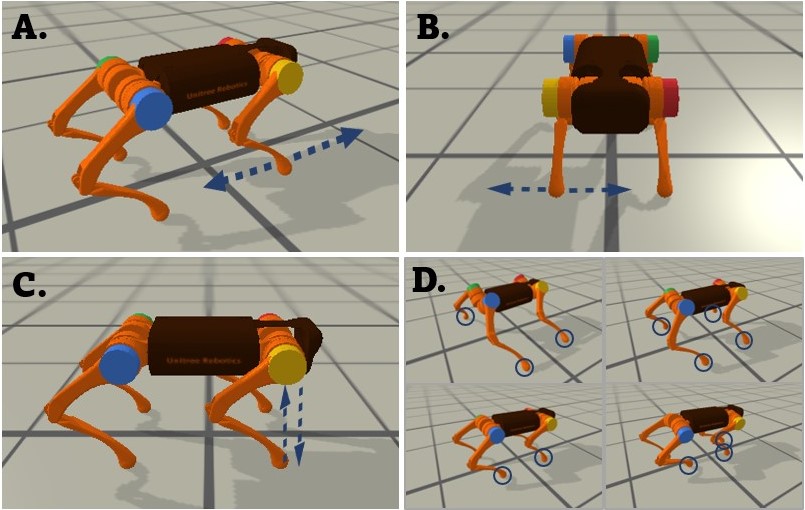}
\centering
\caption{\textbf{Visualisations of the trajectory generator parameters.} \textbf{(A)} Swing - legs move forwards/backwards (along x-axis). \textbf{(B)} Turn - legs move leftwards/rightwards (along y-axis). \textbf{(C)} Lift - legs move upwards/downwards (along z-axis). \textbf{(D)} Y-position of the legs.}
\label{fig:tg_parameters}
\vspace{-4mm}
\end{figure}

The TG determines the relative \(x, y, z\) coordinates of the foot positions for each leg $i$ from the corresponding hip reference frame.
The foot positions are computed based on the periodic phase variable \(\phi\) and the residual TG frequencies \freq~output by the residual policy network. 
The periodic phase $\phi$ which governs the frequency of the gait, ranges from \([0, 2\pi]\) and when \(\phi \in [0, \pi]\) the leg is in swing movement, and when \(\phi \in [\pi, 2\pi]\) the leg is in stance movement.
We use a fixed base frequency $\freq_0$ of 1.25Hz.
The motion of the TG is parametrized by 5 variables which make up the solution vector $\tgparams = [s, t, l, y, \vec{\phi}_0]$ for \algoname. All legs share the same parameters and these are defined as follows:

\begin{itemize}[leftmargin=*]
    \item Swing $s$: determines the maximum forwards and backwards range of the trajectory (Fig. \ref{fig:tg_parameters}A), and the parameter is continuous between 0m and 0.08m from the initial \(x\) position of each leg. 
    \item Turn $t$: determines the maximum inwards (towards the center of the body) or outwards (Fig. \ref{fig:tg_parameters}B), and the range is continuous between 0m and 0.15m from the initial \(y\) position of each leg. 
    \item Lift $l$: determines maximum height foot is raised above the ground (Fig. \ref{fig:tg_parameters}C) during the trajectory, and the range is continuous between 0m and 0.2m from the initial \(z\) position of each leg.
    \item \(y\) position of all legs: determines the initial \(y\) position of each leg (i.e., a negative \(y\) value means the legs are closer to the center of the body and a positive \(y\) value means the legs are further apart from the center of the body) (Fig. \ref{fig:tg_parameters}D), and the range is continuous between -0.05m to 0.12m which is offset to the initial \(y\) position of each leg.
    \item Phase offset $\vec{\phi}_0$: is a vector composed of 4 different phase offset values (i.e. one for each leg) which define the gait coordination between the legs.  
    This parameter can be one of four different gaits: walk \([0, 0.25, 0.5, 0.75\), trot \([0, 0.5, 0.5, 0]\), bound \([0, 0.5, 0, 0.5]\), and pronk \([0, 0, 0, 0]\) and thus, the parameter takes the form of a  discrete integer between 0 and 3.
\end{itemize}

\textbf{\algonamefull}. 
As mentioned earlier, we formulate the problem of finding a diverse set of specialised priors as a QD problem.
QD algorithms solve this problem by maintaining a population of solutions each defined by its performance when evaluated and a descriptor, a vector used to differentiate solutions to encourage diversity. 
We use MAP-Elites (ME), a popular QD algorithm, which tessellates the descriptor space into a grid of cells~\cite{mouret2015illuminating, cully2015robots}.
Because we are interested in solutions that work well in a diversity of environments, the cells of the descriptor space each represent an environment rather than a behavior of a solution. 
Additionally, these are defined beforehand like the multi-task ME setting~\cite{mouret2020quality}.
In other words, we maintain container in the form of grid of diverse environments where each cell in the grid represents a task defined by the environment type and parameterisation (Fig. ~\ref{fig:method_framework} A) and stores the best corresponding solution.
Further details regarding environment parameterisation are provided in Section~\ref{subsec:exp_setup}
Fig.~\ref{fig:method_framework} and Alg.~\ref{algo:map_elites} shows the procedure of \algoname.
\algoname~first begins with random initialisation of solutions in 10\% of the cells in the grid by uniformly sampling from the parameter space target environment in the grid.
Solutions are added in to the corresponding cells even if they have a low fitness.
\algoname~ then follows a loop.
At each iteration, a solution from a cell is first selected. It uniformly selects an existing solution from the grid.
Next, a target cell environment is selected from the grid. 
To encourage effective \textit{goal-switching}~\cite{nguyen2016understanding, clune2019ai}, the target environment type is selected with a $p=0.7$ probability to be same environment type as the solution selected and with $p=0.1$  each of the three remaining environment types.
The environment parameterisation is then sampled uniformly from the range of the corresponding selected environment type.
Effective goal switching is key to \algoname~ and is what makes QD algorithms work well. 
It allows solutions from different environments to act as useful \textit{stepping stones}~\cite{wang2019poet} in the optimization procedure.

The selected solution is mutated/perturbed via a variation operator. We use the iso-line directional variation operator~\cite{vassiliades2018discovering}.
The solution is then evaluated in the target environment by rolling out the open-loop TG motion to obtain the reward and performance of the solution.
If the target environment cell is empty, the solution is inserted in the cell. 
However, if the cell already contains a solution, the fitness scores are compared, and the solution with the highest fitness score is kept in the cell.
This procedure is repeated until we hit the evaluation budget defined.

\subsection{Robustness via Policy Modulation} \label{method_B}
Given a diverse set of TGs, \algoname~builds on the PMTG~\cite{iscen2018policies} control architecture by learning a residual policy represented as a deep neural network to modulate and provide robustness to the base TG using RL.

Fig. \ref{fig:method_framework}B shows an overview of the PMTG control architecture used. 
At each timestep, the policy receives the robot state observation $s_t$, the TG foot frequencies, and the TG swing $s_i$, turn $t_i$ and lift $l_i$ parameters as input. 
The residual policy outputs the \(x_i, y_i, z_i\) residuals for each foot $i$ and the four frequency multiplier $\freq_i$ (one for each leg) to modulate the internal phase of the TGs (see Eqn\ref{eg:ftg}).
The TG then takes TG parameters \tgparams~along with the residual frequencies $\freq_i$ as input to compute the nominal \(x, y, z\) foot positions (see Eqn.~\ref{eg:ftg}). 
The four legs are decoupled for greater flexibility as the frequency $\freq_i$ of each leg can be separately adjusted which leads to more robust locomotion. 
The nominal positions are then summed with the residuals output by the policy to determine the target foot positions.
Inverse kinematics is then applied to the target foot positions to compute the target joint angles which are given to a PD impedance controller used to determine the required torque for each joint.
The policy and TG operate at 60 Hz while the tracking of the target foot positions by inverse kinematics and the PD controller operate at 240 Hz.

\begin{algorithm} 
\caption{\algoname} \label{algo:map_elites}
\begin{algorithmic}
    \Require
    \State $M \gets \text{number of iterations of MAP-Elites (ME)}$
    \State $G \gets \text{ME grid containing environments and solutions}$
    
    \State $G.init()$
    
    \For{$gen = 1, ..., M$} 
        \State $\tgparams, E \gets SELECT(G)$
        \State $\tilde{\tgparams} \gets MUTATE(\tgparams)$
        \State $fitness \gets EVALUATE(\tilde{\tgparams}, E)$
        \State $G \gets ADD(\tilde{\tgparams}, E, fit, G)$
    \EndFor
\end{algorithmic}
\end{algorithm}

% difference with normal PMTG
The residual policy is optimized using Evolutionary Strategies, more specifically Augmented Random Search (ARS)~\cite{mania2018simple} as also done originally by Iscen et al.~\cite{iscen2018policies}.
Conventionally, the policy is trained with only a single pre-defined TG.
In our work, in order to maintain just a single residual policy to handle the diversity of learnt TGs generated, we also condition the residual policy on the TG parameters $\pi(.|\tgparams,s)$.
At each iteration of ARS, we uniformly sample from an environment and its corresponding TG solution from a cell.
To learn robust policies, we add gaussian noise ($\sigma \sim$ discretization of one cell) to the environment parameterization during policy optimization.

\section{EXPERIMENTS} \label{sec:exp}

\subsection{Experimental Setup} \label{subsec:exp_setup}
We evaluate our method on the task of learning to walk forward over a variety of challenging terrains.
We use the Unitree A1 quadruped robot and the PyBullet~\cite{coumans2016pybullet} simulator for all our experiments.
Implementation of TGs and environment setup was adapted from open-source repositories~\cite{coumans2021_tg_github, davide2021github}.
%state space
The state of the robot consists of the CoM base orientation, CoM base velocity, CoM base angular velocity, joint positions, joint velocities and the foot positions in the base frame.
We do not consider any privileged information, which are often used to train robust policies in simulation~\cite{lee2020learning, kumar2021rma}.

\textbf{Environment Setup.}
For our experiments, we consider 80 different environments which make up our environmental grid (as explained in Section~\ref{method_A}). The grid is 4x20 grid composed of four different environment types (Fig~\ref{fig:feature_fig}); slopes, stairs, uneven terrain and balance beams, where we consider a range of 20 environment encoding variations.
The slope environment is parameterised by a steady decline and incline angle that ranges from -11.5\degree to 11.5\degree, where negative values represent walking down the slope and positive values walking up the incline. 
The stairs environment is parameterised by the step height of the stairs ranging from -0.1m to 0.1m where again both descending and ascending stair environments correspond to positive and negative values of the parameters repsectively.
Next, the uneven terrain environment is parameterised by a maximum height variable which varies from -0.1m to 0.1m. The terrain is then constructed by a composition of squares with a fixed size and a random height that is sampled from a uniform distribution in the range from zero to this maximum height parameter. Negative values correspond to holes while positive values correspond to bumps.
Lastly, the balance beam environments are parameterised simply by the width of the beam which varies from 0.25m to 0.75m.

\textbf{Reward Function.} 
We use the same reward function to evaluate the performance of the solution in the EETG phase and the policy training phase.
The reward term \(r_t\) at each timestep is given by Eqn.~\ref{eq:fit_func} and is composed of five terms: target linear and angular velocity tracking reward $r_{lv}$ and $r_{av_t}$, angular velocity penalty $r_{av_p}$, foot position smoothness reward $r_s$, and a, torque penalty $r_{tp}$. 
% The coefficients of each of these terms defines the weight of each term on the final reward \(r_t\).
The performance (also known as the fitness) of a solution is simply the return $R$ which is computed as the sum of the rewards $\sum_{t=0}^T r_t$ obtained at each timestep \(t\) of the episode length $T$.
\begin{equation}
    r_t = r_{lv} + r_{av_t} + r_{av_p} + r_s + r_{tp}
    \label{eq:fit_func}
\end{equation}

\def\arraystretch{1.2}
\begin{table*}[] 
\caption{Summary of \algoname~, baselines, evaluation budget and performance.} \label{table:evals}
\centering
\begin{tabular}{|l|l|l|c|c|c|c|c|c|c|} 
\cline{2-10} \multicolumn{1}{l|}{} & \multicolumn{2}{c|}{\textbf{Alg. Description}} & \multicolumn{3}{c|}{\textbf{Evaluation Budget}} & \multicolumn{4}{c|}{\textbf{Median Total Reward}} \\ \hline

\textbf{Algorithm} & \textbf{TG} & \textbf{Policy} & \textbf{TG Opt.} & \textbf{Policy Opt.} & \multicolumn{1}{c|}{\textbf{Total}} & \textbf{Slope} & \textbf{Stairs} & \textbf{\begin{tabular}[c]{@{}c@{}}Uneven \\ Terrain\end{tabular}} & \textbf{\begin{tabular}[c]{@{}c@{}}Balance \\ Beam\end{tabular}} \\ \hline

\multicolumn{1}{|l|}{\textbf{\algoname}} & Set, Learnt & Single  &  384,000 &  4,608,000  & 4,992,000 & \textbf{194.13} & \textbf{172.73} & \textbf{168.45} & 139.90 \\ \hline
\multicolumn{1}{|l|}{\pmtgvone} & Single, Fixed & Single & -    &  5,280,000 & 5,280,000 & 156.62 & 156.71 & 156.68 & 156.60  \\ \hline
\multicolumn{1}{|l|}{\pmtgvtwo} & Set, Fixed & Set & -   & 23,040,000  &  23,040,000 & 145.35 & 143.13 & 153.17 & 153.05 \\ \hline
\multicolumn{1}{|l|}{\cmaespmtgvone} & Single, Learnt & Single & 384,000  &   4,608,000    & 4,992,000 &  186.85 & 164.19 & 167.79 & 158.60 \\ \hline
\multicolumn{1}{|l|}{\cmaespmtgvtwo} & Set, Learnt & Set & 1,280,000 & 23,040,000  & 24,320,000 & 187.08 & 169.12 & 168.11 & \textbf{165.60} \\ \hline
\end{tabular}
\end{table*}

% Baselines
\textbf{Baselines.} 
We compare our method against two baseline PMTG algorithms to extensively understand and demonstrate the benefits of learning a diverse set of specialized priors in~\algoname.
The first baseline algorithm is the vanilla PMTG algorithm in which a residual policy operates on a single fixed pre-defined TG.
Next, we compare against a baseline in which we first learn a good TG using the Covariance Matrix Adaptation Evolution Strategy (CMAES)~\cite{hansen2003reducing} and consequently train the residual policy over this learnt TG. We refer to the this baseline as CMAES-PMTG.
Both baselines use the same policy and PMTG architecture and use ARS for the policy optimization stage.

For a more complete comparison, we consider two different variations of the baseline vanilla PMTG and CMAES-PMTG algorithms:

\begin{itemize}[leftmargin=*]
    \item \textbf{Encoded (Enc.):} 
    The first variant considers only a single TG and a single residual policy that is optimised on all the environment types and variations. 
    To enable the policy to adapt to the environments, the policy receives the environment type and variation as a one-hot encoded vector input.
    \item \textbf{Independent (Ind.):} This variation is used as an upper baseline. We use independent TGs and policies for each environment (i.e. one TG and one residual policy per environment cell) in the grid. 
    As a result, each TG and policy can specialise for each specific environment.
    This variant requires significantly more evaluations and is the least sample efficient as we have to independently train TGs (CMAES-PMTG baseline) and policies (both baselines) in each environment.
\end{itemize}

The total evaluation budget for training given to each variant is provided in Table~\ref{table:evals}.
% Metrics
To evaluate each algorithm, we take the final TG(s) and optimized residual policies provided at the end of each algorithm. 
This is deployed in each of the 80 environments in the grid. 
To ensure that we evaluate robustness to unseen or variations in the environment, sampled gaussian noise is added to the environment parameterization during each evaluation.
As a metric, we report the average total reward obtained for each environment over 20 replications of evaluations.

\subsection{Results}
\begin{figure}[t]
\centering
	\includegraphics[width=0.48\textwidth]{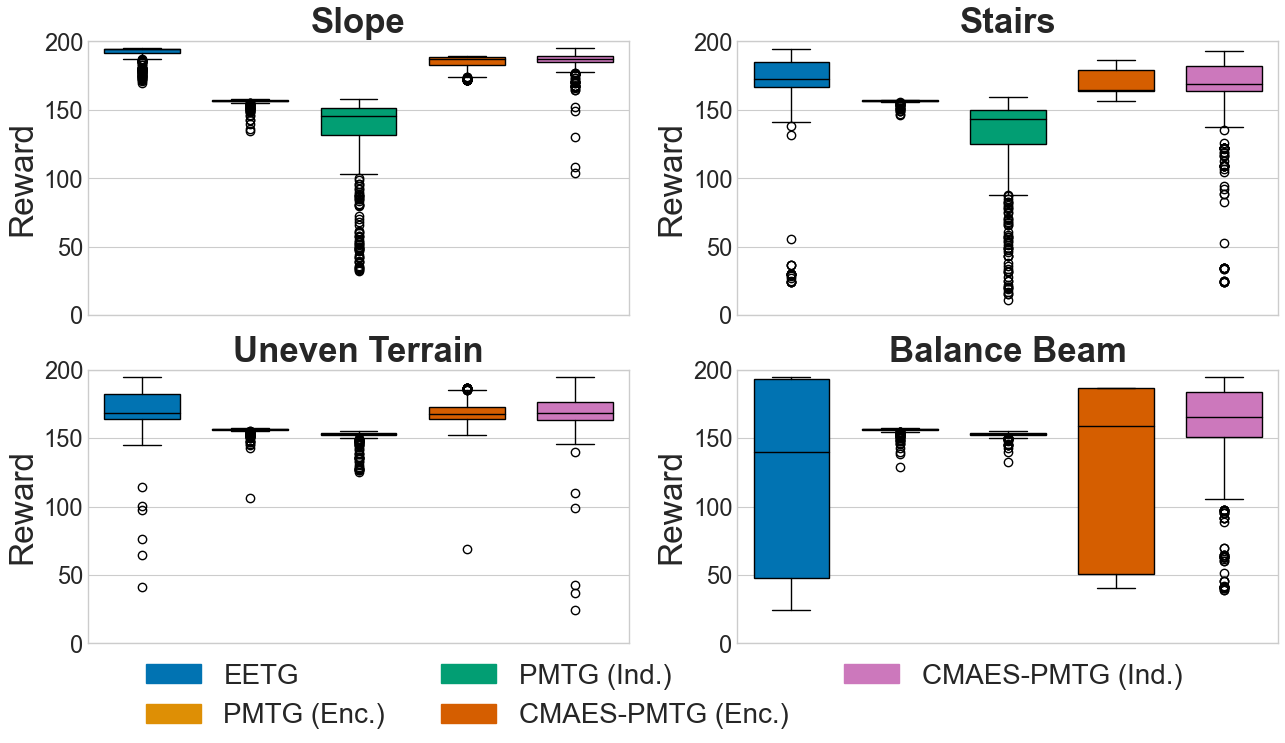}
\centering
\caption{The total reward obtained by \algoname~(blue) and the baselines: \pmtgvone~(orange), \pmtgvtwo~(green), \cmaespmtgvone~(red), and \cmaespmtgvtwo~(pink). The results are shown separately for each environment type.}
\label{fig:main_results}
\vspace{-4mm}
\end{figure}

Figure \ref{fig:main_results} and Table~\ref{table:evals} shows the results of the performance of \algoname~and baselines for each environment type.
Despite using the least number of samples, we can observe that \algoname~performs the best and is capable of traversing a variety of diverse terrains (see accompanying video). 
This can be attributed to a few factors explained by the performance of the baselines.
First, we can observe that both vanilla PMTG variants obtained the lowest performance across all the environment types when compared to both \algoname~and CMAES-PMTG baselines.
This result demonstrates the importance of a learnt TG prior over the constraints and biases of a fixed TG prior.
This is because a TG that is learnt, already begins with a higher performance than a pre-defined TG. 
Thus, this shows that the samples used to learn a good TG prior on the task is more effective for learning than providing more samples to train policy with a fixed TG prior. 
%does not require a significant amount of samples to further improve performance.

Next, we can also observe that the algorithms with a set of specialized TG's such as \algoname~and \cmaespmtgvtwo~tend to perform better than the variants with only a single TG.
This observation can be explained by our earlier hypothesis where a single prior can struggle when learning on a diverse set of tasks/environments and reduces its effect in helping the learning algorithms.
Learning a diverse set of TG priors can help to alleviate this when dealing with a diversity of environments and restore the effectiveness of the prior.
Additionally, as observed previously, the process of learning a prior seems to be more sample efficient than learning a policy directly which makes learning a set of TG priors feasible.

Lastly, but also most importantly, we can observe that \algoname~performs slightly better or very similarly to \cmaespmtgvtwo~ despite requiring five times less samples.
This demonstrates that EETG can learn locomotion skills efficiently across a diversity of environments as good or better than a complete set of independently trained policies and TGs.
We hypothesize that this is due to the goal switching phenomena and the availability of useful stepping stones in QD algorithms that we use in \algoname. 
This also corresponds to observations in literature where it was shown that learning occurs more efficiently through the process of goal switching across a diversity of tasks in comparison to the more brute force goal directed optimization~\cite{lehman2010efficiently, woolley2011deleterious}.

One additional observation is that all algorithms (including \algoname) perform badly on the balance beam environment. 
This can be explained by the fact that our robot is blind and only considers proprioception while this environment can easily result in catastrophic failure (i.e falling off) and would require vision information to correct behaviors.
This is in contrast to the other three environments in which proprioception can be used to react effectively.

\subsection{Ablation}
To further evaluate and find the optimal configuration for \algoname, we aim to answer the following questions by running an ablation against a few variations of \algoname:
(1) Does adding repeated, sequential optimisation of the MAP-Elites grid and residual policy help performance?
%when compared with \algoname~that has one MAP-Elites loop followed by one PMTG loop? 
(2) Does including the policy during the MAP-Elites loop help performance? 

We investigate this by comparing against two variants of \algoname~, both of which consists of multiple iteration loops between the two optimization phases: MAP-Elites and Policy Optimization.
In \algovarone, the policy is not used in the MAP-Elites TG learning stage where only the open loop TG is used.
In \algovartwo, the latest policy from the policy optimization phase is also used in the controller during the MAP-Elites TG learning stage.
In this experiment, all variants are given the same total evaluation budget as the default \algoname~ but divided equally between the number of loop iterations.

\begin{figure}[t]
\centering
	\includegraphics[width=0.48\textwidth]{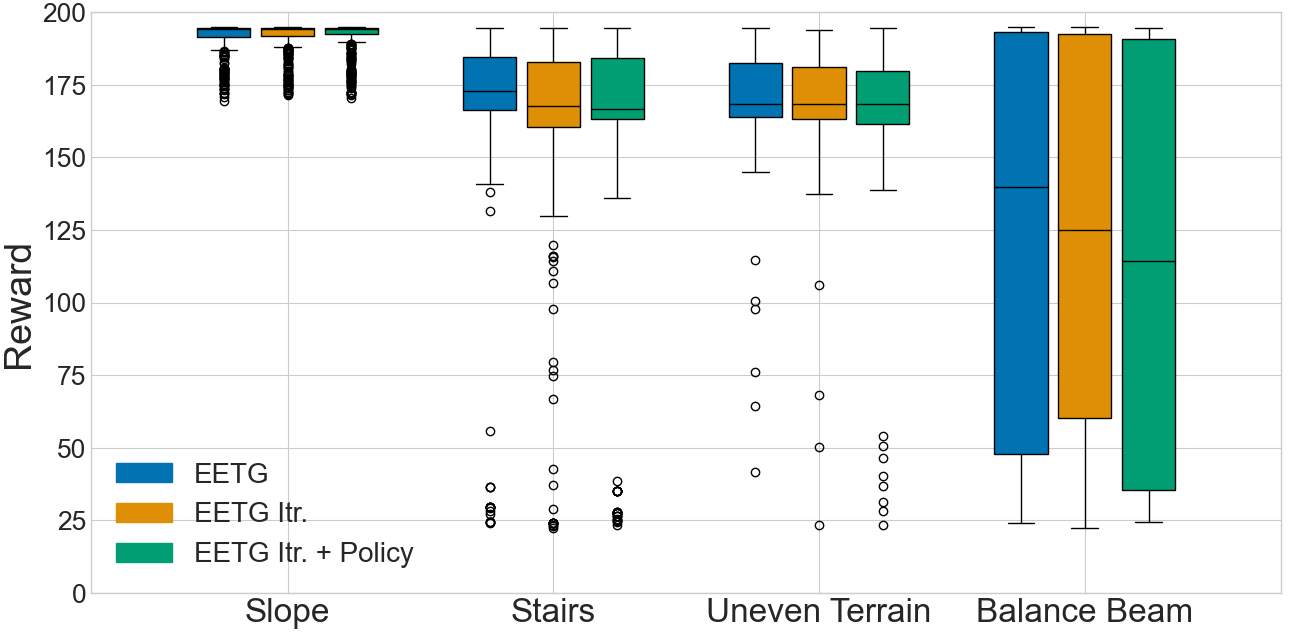}
\centering
\caption{The total rewards obtained by \algoname~(blue), and the two variants, \algovarone~(orange) and \algovartwo~(green). The results are shown separately for each different environment types.}
\vspace{-4mm}
\label{fig:ablation_results}
\end{figure}

Figure \ref{fig:ablation_results} shows the performances of \algoname~and its variations, \algovarone~and \algovartwo, for each environment type. 
It can be seen that \algoname~obtains relatively better rewards across all the environments. 
However, the difference in total rewards between \algoname~and its variants only ranges from 0\% to 4\% of the maximum total reward that can be obtained in one episode.
Furthermore, the differences in the observed behaviours of the robot trained with each algorithm are minimal. 
Therefore, it cannot be concluded that these variants lead to significant changes in the final performance when compared to \algoname. 
Based on these results, it does not seem that (1) including multiple iterations of the ME and PMTG loop or (2) involving the residual policy in the ME evaluation stage lead to any significant changes in the final performance of the \algoname~algorithm.

\section{DISCUSSION AND CONCLUSION}
% Summary
In summary, we present \algoname, a novel method which evolves a set of high-performing TG priors for a diversity of environments and uses a single residual policy that regulates the learnt TGs.
We show that this enables a quadruped robot to traverse a wide range of challenging environments. 
Our experiments highlight the need and effectiveness of a set of specialised TG priors over a single fixed manually defined TG prior when dealing with a variety of environments.
We demonstrate that ~\algoname~ is significantly more sample efficient and as effective as learning independent TGs and policies across diverse environments.
%than the strongest and most competitive baseline, \cmaespmtgvtwo. 

% Limitations and Future work
% balance beam - vision
One of the limitations of this work is that we assume knowledge of what TG in the set to use during deployment. In future work, it would be interesting to eliminate this assumption in order to autonomously and efficiently select a suitable TG using Bayesian Optimisation through Intelligent Trial and Error~\cite{cully2015robots, chatzilygeroudis2018reset} via online data collected during deployment.

\addtolength{\textheight}{-1cm}   % This command serves to balance the column lengths

%%%%%%%%%%%%%%%%%%%%%%%%%%%%%%%%%%%%%%%%%%%%%%%%%%%%%%%%%%%%%%%%%%%%%%%%%%%%%%%%

%%%%%%%%%%%%%%%%%%%%%%%%%%%%%%%%%%%%%%%%%%%%%%%%%%%%%%%%%%%%%%%%%%%%%%%%%%%%%%%%


\section*{ACKNOWLEDGMENT}
We thank Manon Flageat for reviewing earlier versions of this paper. This work was supported by the Engineering and Physical Sciences Research Council (EPSRC) grant EP/V006673/1 project REcoVER.

%%%%%%%%%%%%%%%%%%%%%%%%%%%%%%%%%%%%%%%%%%%%%%%%%%%%%%%%%%%%%%%%%%%%%%%%%%%%%%%%

\bibliographystyle{IEEEtran} % use IEEEtran.bst style
\bibliography{biblio}

\end{document}